Citation: Akogo DA and Palmer XL (2018) ScaffoldNet: Detecting and Classifying Biomedical Polymer-Based Scaffolds via a Convolutional Neural Network ArXiv

Uploaded: May 16, 2018

Funding: No addtional external funding was used to produce this work

RESEARCH ARTICLE

# ScaffoldNet: Detecting and Classifying Biomedical Polymer-Based Scaffolds via a Convolutional Neural Network

**Darlington A. Akogo[1], Xavier-Lewis Palmer[1,2]**

1  minoHealth  AI  Labs,  00233,  SCC  Inside,  Weija,  Accra,  Ghana
2  Biomedical  Engineering  Institute,  Old  Dominion  University,  Norfolk,  VA  23529,
United States of America

E-mail: darlington@gudra-studio.com

Keywords: Polymer, Detection, Biocompatible, Nanofibers, AI, Neural Network, Machine Learning, Convolutional, Biomimetic, 3D, Fiber, Biology, Screening

## Abstract

We developed a Convolutional Neural Network model to identify and classify Airbrushed (alternatively known as Blow-spun), Electrospun and Steel Wire scaffolds. Our model ScaffoldNet is a 6-layer Convolutional Neural Network trained and tested on 3,043 images of Airbrushed, Electrospun and Steel Wire scaffolds. The model takes in as input an imaged scaffold and then outputs the scaffold type (Airbrushed, Electrospun or Steel Wire) as predicted probabilities for the 3 classes. Our model scored a 99.44% Accuracy, demonstrating potential for adaptation to investigating and solving complex machine learning problems aimed at abstract spatial contexts, or in screening complex, biological, fibrous structures seen in cortical bone and fibrous shells.

## Introduction

Convolutional Neural Networks have been an important aspect of Deep Learning in recent years. They were mainly responsible for the re-emergence and popularity of Neural Networks. The work of Alex Krizhevsky and Ilya Sutskever which won the ImageNet Large Scale Visual Recognition Competition in 2012 (ILSVRC-2012) was disruptive in the Artificial Intelligence, Machine Learning and Computer Vision community (Alex Krizhevsky et al., 2012). Since then Convolutional Neural Networks have been heavily applied to all sorts of problems, from various Object Detection and Image Segmentation problems (Liang-Chieh Chen et al., 2014, J. Redmon et al., 2015, S. Ren et al., 2015) and to specific domains like Medical Image Analysis (Shadi Albarqouni et al., 2016, Mark J. J. P. van Grinsven et al., 2016, Lin Yang et al., 2017, Andre Esteva et al., 2017).

Neural Networks including Convolutional Neural Networks were unfortunately generally abandoned in the late 1980s. The reason why they were abandoned and why they re-emerged can both be attributed to Computational Power and Amount of Data. Deep Neural Networks require a lot of processing power to be effectively trained and they only perform well when trained on a lot of Data. Both were lacking then and have grown a lot in recent years. With lots of powerful GPUs and Big Data, Neural Networks finally could be applied to complex problems.

Convolutional Neural Networks are very effective for Computer Vision problems. After winning the ILSVRC-2012 Competition, Convolutional Neural Networks have been applied to multitudes of Computer Vision problems. Their effectiveness can be attributed to their ability to handle translation invariances in images by relying on shared weights and exploit spatial locality by enforcing a local connectivity pattern between neurons of adjacent layers. We chose them for this reason, knowing we wanted a model that could visually detect and differentiate between different types of scaffolds.

In this particular example, we utilized our convolutional neural network towards demonstrating that they can be used for distinguishing between different scanning electron microscope images of polymer-based scaffolds manufactured for research and application tissue regeneration. Three scaffold types were used. One was from an airbrushed set, and two others were from training study (Hotaling et al 2016), composed of an electrospun fiber set and a control set composed of steel wires. We set out to test if we could develop a Convolutional Neural Network model that could identify and classify among the sets in hopes of producing a tool that would be useful in biomedical manufacturing, forensics, and perhaps more.

## Problem Formulation

Our problem was framed as a Classification problem, given a 128 x 128 pixels image of a scaffold, our model has to classify it as either Airbrushed, Electrospun or Steel Wire. Our model's objective during training is to optimize the Cross-entropy loss

$$-\sum_{c=1}^{M} y_{o,c} \log(p_{o,c})$$

where **M** - number of classes (3: Airbrushed, Electrospun and Steel Wire)
log - the natural log
**y** - binary indicator (0 or 1) if class label c is the correct classification for observation o
**p** - predicted probability observation o is of class c

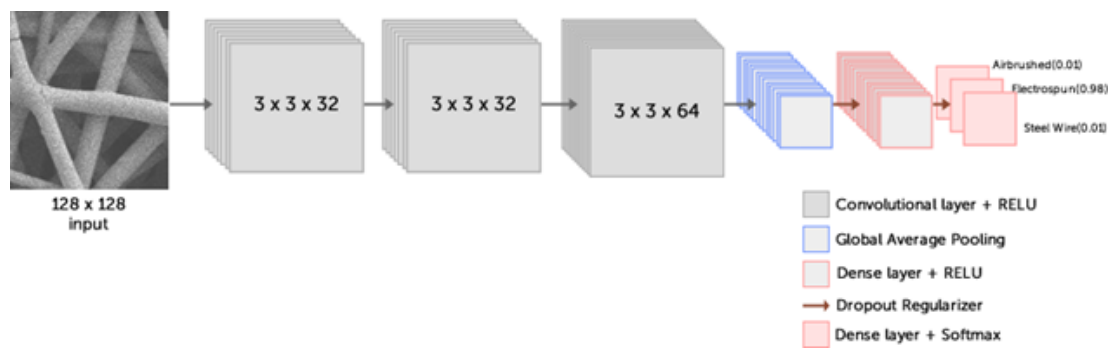

**Figure 1.** The architecture of our Convolutional Neural Network: ScaffoldNet. This shows a graphical demonstration of ScaffoldNet at work, it takes a 128 x 128 grayscale image of an Electrospun scaffold in its input layer (the 1st 3x3x32 Convolutional layer), propagates it through all its 4 hidden layers then finally outputs predicted probabilities for the 3 classes in its 3 unit output layer (the last Dense layer). It outputs the following probabilities for the 3 classes respectively; Airbrushed (0.01), Electrospun (0.98) and Steel Wire (0.01). ScaffoldNet accurately identified and classified the scaffold image as Electrospun by assigning it the highest probability among the classes (Electrospun: 0.98).

## Model Architecture: Convolutional Neural Network

We used a 6-layer Convolutional Neural Network trained on our 3,043 scaffold images dataset. As shown in Figure 1, our network ScaffoldNet starts with two 2-Dimensional Convolutional layers with a 3 x 3 kernel size and 32 output filters with the first as its input layer. Then followed by a single 2-Dimensional Convolutional layers also with a 3 x 3 kernel size and 64 output filters. The receptive fields of the Convolutional layers' filters (equivalently this is the filter size) captures various features across local regions in our input images, which is why they are powerful. During the forward propagation, our network slides ( convolves) each filter across the width and height of the input images and computes dot products between the entries of the filter and the input at any position. As our network slides the filter over the width and height of the input images, it will produce a 2-dimensional activation map that gives the responses of that filter at every spatial position. During training, our model network will learn filters that activate when they see some type of visual element such as an edge of some orientation or a blotch of some color on the first layer, or eventually entire honeycomb or wheel-like patterns on higher layers of the network.

We then introduce a 2-Dimensional Global Average Pooling layer to reduce the spatial dimensions of our tensor (Min Lin et al., 2013). Global Average Pooling performs Dimensionality Reduction to minimize overfitting by turning a tensor with dimensions **h × w × d** into 1 × 1 × d which is achieved by reducing each **h × w** feature map to a single number simply by taking the average of all **hw** values. To further prevent overfitting, we then add a Dropout Regularizer with a fraction rate of 0.5 (Srivastava et al., 2014) which is a Regularization technique that prevent Neural Networks developing complex co-adaptation on the training data. We then introduce a 32 unit densely-connected Neural Network layer into our network architecture, followed by another Dropout Regularizer with a 0.5 fraction rate. Our final output layer is 3 unit densely-connected Neural Network layer.

All Convolutional and Densely-connected layers except the output layer use the Rectified Linear Unit (RELU) activation function:

$$f(x) = max\ (0, x)$$

where **x** is the input to a neuron (Hahnloser et al., 2000).

RELUs are currently the most popular and successful activation function because they allow Deep Neural Networks to be more easily optimized than sigmoid and tanh activation functions due to the fact that gradients are able to flow when the input to the ReLU function is positive (Ramachandran et al., 2017). Our final densely-connected output layer uses a Softmax activation function:

$$\sigma(z)_j = \frac{e^{z_j}}{\sum_{k=1}^{K} e^{z_k}}$$

where z is a vector of the inputs to the output layer (we have 3 output units, so there are 3 elements in z). And again, j indexes the output units.

Softmax function squashes the raw class scores into normalized positive values, then outputs them as separate probabilities for each of our classes(Airbrushed, Electrospun and Steel Wire), where all the probabilities add up to 1.

ScaffoldNet is trained end-to-end with Adam optimization algorithm using the standard parameters (β1 = 0:9 and β2 = 0:999) (Kingma and Ba, 2014). Adam is an algorithm for first-order gradient-based optimization of stochastic objective functions which computes individual adaptive learning rates for different parameters from estimates of first and second moments of the gradients. We train our model using mini-batches of 32. We use a learning rate (α) of 0.001, and pick the model with the lowest validation loss.

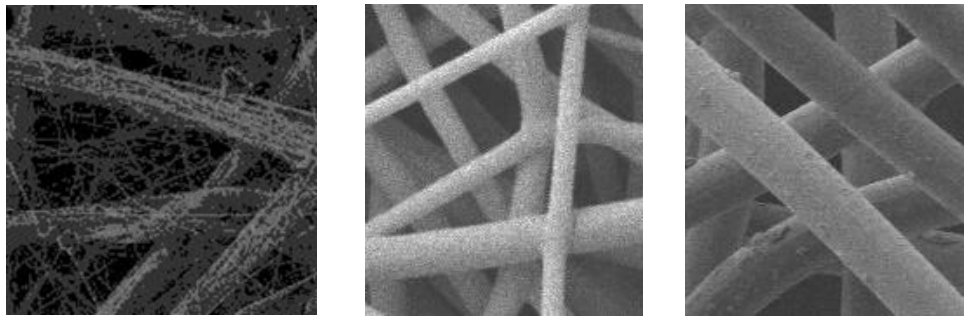

(a) Airbrushed sample      (b) Electrospun sample      (c) Steel Wire sample

**Figure 2.** Samples of the imaged Scaffolding materials belonging to the 3 classes (Airbrushed, Electrospun and Steel Wire) used in training ScaffoldNet.

## Data

The dataset used for training, validating and testing our model is a collection of 3,043 grayscale images of Airbrushed, Electrospun and Steel Wire scaffolds. We can see sample images from the 3 classes in Figure 2 above. The dataset contains 1013 Airbrushed scaffold images, 960 Electrospun scaffold images and 1070 Steel Wire scaffold images as shown in Figure 3. The airbrushed dataset is derived from scanning electron microscope images of airbrushed scaffolds under multiple polymer admixture settings taken at 200x magnification via a JEOL 1990 scanning electron microscope. Details regarding the electrospun and steel wire set can be found Hotaling et al 2015. In total, the three classes are Airbrushed, Electrospun, and Steel Wire. Each of the sets are SEM images taken at 200x magnification, giving them suitable plane for comparison.

The images of each class differ largely with respect to two factors, porosity and fiber diameter. In terms of appearance, the airbrushed class is quite easy to distinguish from the electrospun and steel wire class, but this ease disappears when comparing between the electrospun and steel wire classes without prior training. A useful AI with regards to manufacturing and forensic utilities would need to be able to differentiate between scaffold classes, ideally at extremely high rates for bulk processing. The impact of such a model is wide in terms of machine optimization and analysis in bulk studies. As per Hotaling et al 2015 and 2016, processes capable of batch analysis can vastly improve workflow and the pace of nanofiber scaffold research.

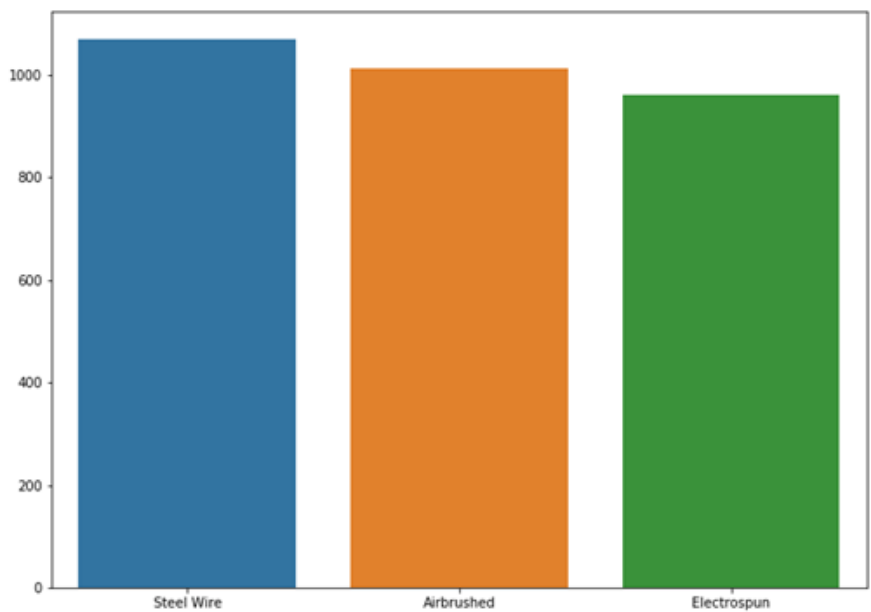

**Figure 3.** The dataset contains 1070 Steel Wire scaffold images, 1013 Airbrushed scaffold images and 960 Electrospun scaffold images.

**Figure 3.** The dataset contains 1070 Steel Wire scaffold images, 1013 Airbrushed scaffold images and 960 Electrospun scaffold images.

## Data Preparation

The collected set of images was then prepared and divided into subsets to best suit Machine Learning. The dataset was split into Training set, Validation set and Testing set with a 8.8 : 1.2 : 1.0 ratio(2376, 368, 301), respectively. The splits were stratified so each Scaffold image class still gets equally split with an 8.8: 1.2: 1.0 ratio between the Training set, Validation set and Testing set. This is to prevent under training on some classes compared to others. The dimensions of all the images were reshaped to 128 x 128 pixels. The images were then transformed by Standardization, which entails setting each image sample's mean to 0 and dividing its pixel values with its Standard Deviation. We do this so our image pixel values that would act as inputs to our model would have a similar range in order to have more stable gradients during training. The images were then further augmented with random horizontal flips, 5° rotations, width shifts, height shifts and zooms. This is to increase the variation in our dataset in order to make our trained model generalize.

## Model Training and Validation

Using the Training set (2376 images), our Convolutional Neural Network was trained with Adam optimization algorithm and Cross-Entropy loss function and the Accuracy Classification Score and Mean Absolute Error(MAE) were used as metrics. The Accuracy score and MAE formulae are as follow respectively;

$$\texttt{accuracy}(y, \hat{y}) = \frac{1}{n_{\text{samples}}} \sum_{i=0}^{n_{\text{samples}}-1} 1(\hat{y}_i = y_i)$$

$$\text{MAE}(y, \hat{y}) = \frac{1}{n_{\text{samples}}} \sum_{i=0}^{n_{\text{samples}}-1} |y_i - \hat{y}_i|.$$

**where the** $\hat{y}_i$ is the predicted output for the **i** -th sample $\hat{y}_i$ is the (correct) target output computed over $n_{\text{samples}}$

ScaffoldNet was trained in 11 epochs and its hyperparameters tuned using the Validation set (368 images). After just the first epoch, our model had the following performance results on the Validation set;

Accuracy score: 64.44%
Cross-Entropy loss: 0.6376
Mean Absolute Error: 0.2803

Our model performance was drastically improving as training continued. After the 11th epoch, our model's final performance results on the Validation set were;

Accuracy score: 100%
Cross-Entropy loss: 0.0100
Mean Absolute Error: 0.0061

## Model Testing and Results

After all 11 epochs of training with the Training set, validation and hyperparameter tuning with the Validation set, our model was finally evaluated on the Test set (368 images). With this final evaluation, we get a better picture on how well our model generalizes and performs on unseen data since the Training set is used during the model's training and the Validation set is used to observe the performance of the model and to tune its hyperparameters to improve such performance. The Test set is the final evaluation for a model and changes are not made to the model after the results.

ScaffoldNet's final performance results on the Test set were;

**Accuracy score: 99.44%**
**Cross-Entropy loss: 0.0997**
**Mean Absolute Error: 0.0101**

From the results of ScaffoldNet's final evaluation, we can tell that our model generalizes well and doesn't overfit, the performance on the Validation set after the 11 epochs and hyper parameter tuning is consistent with the evaluation results on the Test set. Our Convolutional Neural Network consistently demonstrates extremely high accuracy performance beyond even our expectations.

To further evaluate ScaffoldNet's output quality, we use the Receiver Operating Characteristic (ROC) metric and its Area Under Curve (AUC) score. We use the ROC curve to plot our model's true positive rate on the Y axis, and false positive rate on the X axis. ROC curves are mainly for Binary Classification and since our Convolutional Neural Network is Multiclass Classifier, we extended the ROC curve by drawing the ROC curve per class. We also drew another ROC curve by considering each element of the label indicator matrix as a binary prediction (micro-averaging). And we also used macro-averaging, which gives equal weight to the classification of each label. The top left corner of the plot is the "ideal" point - a false positive rate of zero, and a true positive rate of one. The ideal point has an AUC score of 1.00, the closer a classifier's score is, the better the classifier.

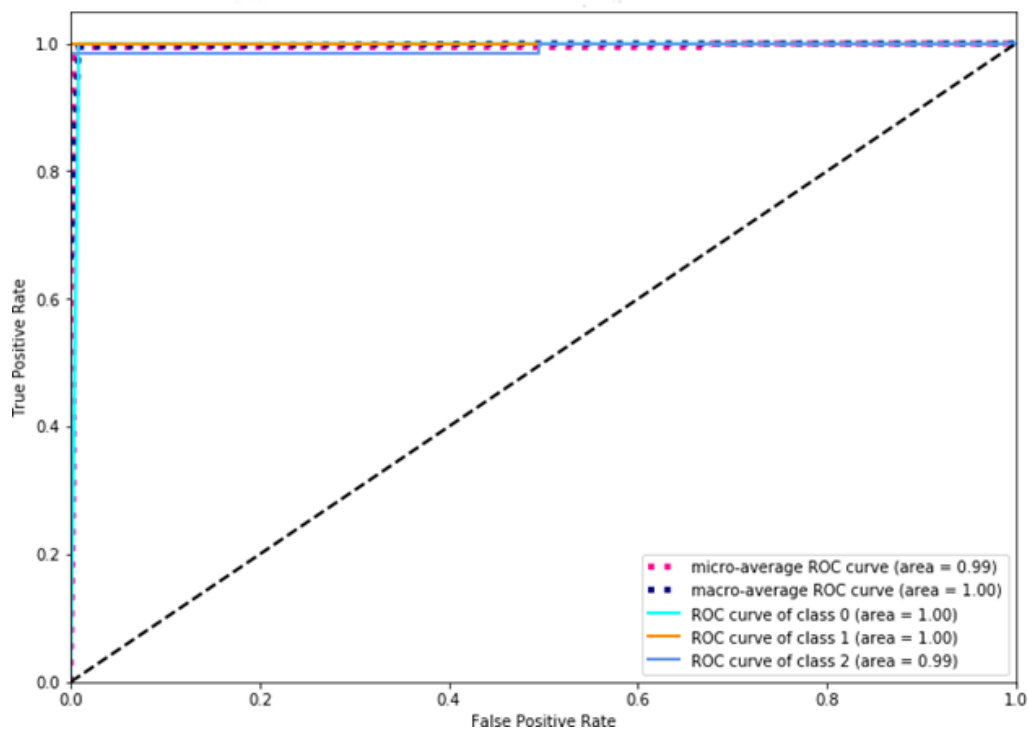

**Figure 4.** Class 0, Class 1 and Class 2 are Airbrushed, Electrospun and Steel Wire classes respectively. As seen on the curve, our Classifier has an extremely optimal performance and score, a micro-average AUC score of 0.99, and a macro-average AUC score of 1.00 which is the highest possible AUC score.

Our Convolutional Neural Network classifier ScaffoldNet has near perfect AUC score as shown in the plot above, both Class 0 (Airbrushed) and Class 1 (Electrospun) have an ideal score of 1.00 and Class 2 (Steel Wire) has a 0.99 score with only a few false positive. The micro-average AUC score of all 3 ROC curves being 0.99 and macro-average AUC score being 1.00. The results from our ROC curve and AUC score are also consistent with our model's evaluation results on the Test set.

The state of the art performance and results of ScaffoldNet across multiple metrics can be explained by the clear and strong visual dissimilarities across our 3 scaffold classes as can be seen in Figure 2. With such visible visual differences between our classes, classification especially with a powerful algorithm like Convolutional Neural Network becomes an easy task. Our model easily found the right visual features in the form of weights to use in identifying and classifying the 3 scaffold types.

## Regarding Related Works

A close related work can be found in Chen et al 2016 in which machine learning is used to identify cell shape phenotypes that have particular micro-environmental cues. Our paper can be used to address a gap in that we are demonstrating a feasible focus on the extracellular matrix mimicking biocompatible polymer based scaffolding, which would directly affect the microenvironment, which is found to be increasingly important in cell behavior and means of tissue engineering (Sachs et al 2017). Our techniques and algorithms can be seen as an updated follow up in spirit, given the dated techniques and algorithms used in theirs, lacking true computer vision according to current standards. Our Deep Learning model is end to end, with the ability to learn to analyze and classify images on its own without need for excessive manual feature engineering. Their model requires Image Processing and Shape Quantification with Snake algorithm and Branching analysis, followed by "Filtering" Feature Selection and Heterogeneity Reduction before running it through a classic Machine Learning classifier such as a Support Vector Machine, whereas our Deep Neural Network model ScaffoldNet is trained to analyze and classify raw images on its and be able to improve with more training data. Our system uses a Convolutional Neural Network in contrast, which is state of the art in Computer Vision and Image/Object Detection.

## Conclusion and Outlook

We developed a Convolutional Neural Network called ScaffoldNet that accurately classifies Airbrushed, Electrospun, and Steel Wire scaffolds after being trained on 2376 scaffold images, validated with 368 scaffold images and tested on 301 scaffold images. Our model demonstrates state of the art results including 99.44% Accuracy score, 0.0101 Mean Absolute Error, 0.99 micro-average AUC score and a 1.00 macro-average AUC score. We've shown the potentials of using Convolutional Neural Network, Deep Learning and Artificial Intelligence as Classification tools for differentiating between 3D Biomedical Polymer-based scaffolds. Our research is also meant to demonstrate the many possibilities and potentials of combining two of arguably the most revolutionary technologies of time, Artificial Intelligence and Biotechnology. We believe that plenty opportunities exist in which we can combine the two fields, mostly by using Artificial Intelligence to automate and optimize the many processes involved in Biotechnology, especially within automating identification and minimization of defects in manufacturing processes and forensics for material on the microscale. Specifically, we are interested in expanding the useful parameters by which the explored classes can be further distinguished and analyzed by our Convolutional Neural Network. Broadly, we intend to further explore many more of such ways in which we can combine Artificial Intelligence and Biotechnology at different measurement scales.

## Acknowledgment

We would like to thank Dr. Carl Simon Jr. and Dr. Nathan Hotaling of the National Institute of Standards and Technology for the for their electrospun nanofiber and steel wire images that were used for this set. Additionally, we would like to thank Dr. Wojtek Tutak of the Food and Drug Administration's Center for Devices and Radiological Health for his mentorship and insights on this manuscript. Finally, we would like to acknowledge Dietmar W. Hutmacher, the Director of ARC Center in Additive Biomanufacturing and Paul Dalton, Professor in Biofabrication at the University of Wurzburg, for sharing some images of Electrospun scaffolds with us for this research work.